\documentclass[11pt]{article}

\usepackage[final]{acl}

\usepackage{times}
\usepackage{latexsym}
\usepackage[T1]{fontenc}
\usepackage[utf8]{inputenc}
\usepackage{microtype}
\usepackage{inconsolata}
\usepackage{graphicx}
\usepackage{multirow}
\usepackage{amsmath}
\usepackage{amssymb}
\usepackage{dsfont}
\usepackage{booktabs}
\usepackage{hyperref}
\usepackage{makecell}
\usepackage{fancyhdr}

\usepackage{svg}
\usepackage{lscape}

\usepackage{algorithm}
\usepackage{algpseudocode}

\newcommand{\Sim}{\mathrm{sim}}

\title{Cultural Evolution Drives the Emergence of Conventions\\in Neural Emergent Communication}

\author{
 \textbf{Maxime Toquebiau\textsuperscript{1}},
 \textbf{Jérôme Botoko Ekila\textsuperscript{1}},
 \textbf{Paul Van Eecke\textsuperscript{1}}\footnotemark[1],
 \textbf{Katrien Beuls\textsuperscript{2}}\footnotemark[1]
\\
\\
 \textsuperscript{1}Artificial Intelligence Laboratory, Vrije Universiteit Brussel, Belgium \\
 \textsuperscript{2} Faculté d'Informatique, Université de Namur, Belgium 
\\
 \small{
   \textbf{Correspondence:} \href{mailto:maxime.toquebiau@vub.be}{maxime.toquebiau@vub.be}
 }
}

\begin{document}
\maketitle

\begingroup
\renewcommand{\thefootnote}{\fnsymbol{footnote}}
\footnotetext[1]{Joint last authors.}
\endgroup

\pagestyle{fancy}
\fancyhead[L]{\small Published in the \textit{Proceedings of the 2026 Conference on Empirical Methods in Natural Language Processing}}
\fancyhead[R]{}

\begin{abstract}
Recent studies of language evolution with neural agents often use models where agents can either speak or listen, where there are only two agents in the population, or where agents can share internal representations. These choices prevent the study of how language emerges through cultural evolution, i.e. through the cumulative adaptation of individual behaviours to fit the communicative needs of the community. We tackle this issue with a new model that trains populations of bi-directional neural agents in a fully distributed fashion to solve a referential game. We demonstrate the effectiveness of this approach on multiple image datasets and show that it is able to scale to large populations. We also show that more conventional languages are easier to learn, providing a direct advantage for cultural transmission. Overall, this work provides a novel approach to neural emergent communication that opens up new directions for evolutionary linguistics. 
\end{abstract}

\section{Introduction}

Computational models of language evolution have played an insightful role in our understanding of the origins of human language. They allow us to test hypotheses on the roles of different environmental, cognitive or cultural biases in the emergence of natural language~\cite{Cangelosi2002, Kirby2002_ALife, Oudeyer2007, Steels2012_Experiments}. 
This approach is rooted in the view of languages as complex adaptive systems subject to evolutionary dynamics~\cite{GellMann1992_Complex, Smith2003_CompoEmerg}. Languages emerge and evolve to fit the communicative needs of populations of individuals~\cite{Dessalles2007_Why}. This can be modelled with artificial agents playing many interactions of language games~\cite{Steels1997_Grounding}, progressively bootstrapping shared communicative strategies grounded in their observations.

Over the past decade, the field of evolutionary linguistics has seen a new wave of interest with agents modelled as neural networks trained with deep learning techniques~\cite{Kottur2017_Natural, Lazaridou2017_Multi, Kharitonov2019_EGG, Chaabouni2020_Composit}. Neural models of emergent communication allow us to scale language game experiments to more complex environments and languages~\cite{Lazaridou2018_Emergence, Chaabouni2022_Scale}. 
Previous work has used neural models to investigate specific mechanisms that may influence language emergence, including cognitive biases~\cite{Lian2021_Effect, Tucker2022_VQVIB}, contextual information~\cite{Gualdoni2024_Bridging, Kobrock2024_Context} and population structure~\cite{Graesser2019, Rita2022_PopHetero, Michel2023_Revisiting}. 

\begin{figure}[t]
    \centering
    \includegraphics[width=\linewidth]{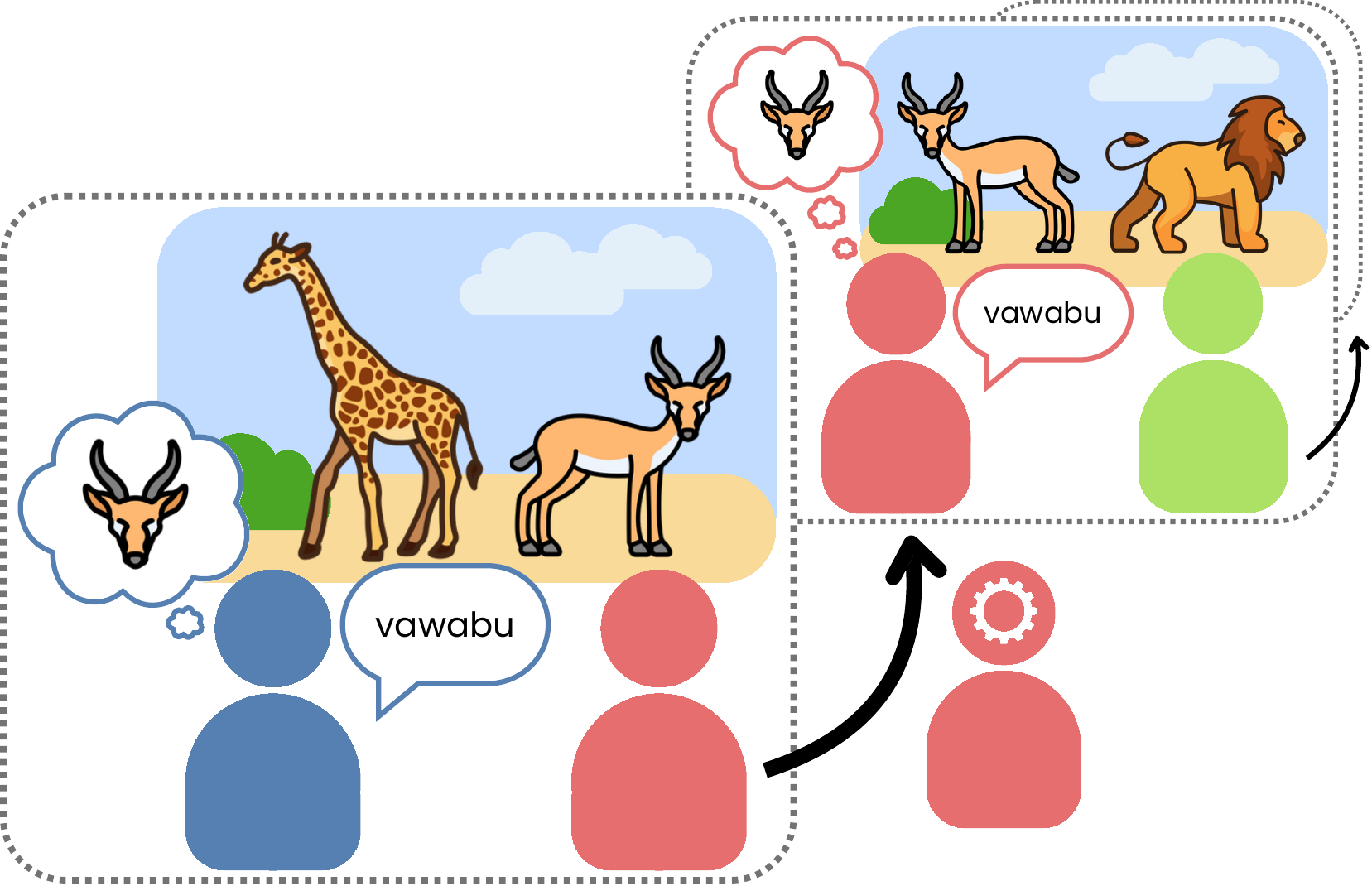}
    \caption{Cultural evolution of linguistic conventions for communicating about the world. Through repeated interactions with members of their community, individuals observe communicative behaviours, internalise them, and reuse them in later exchanges.}
    \label{fig:cultural_evo}
    \vspace{-1.2em}
\end{figure}

We are specifically interested in the mechanisms of cultural evolution that allow the emergence of natural language traits. One especially important trait is conventionality~\cite{Wright2026_Conventionality}. Because linguistic constructions are arbitrary, they must rely on conventions to be of any use~\cite{Clark1996}. These conventions emerge from the need to communicate about the world with various interlocutors. As individuals take part in communicative interactions, communicative strategies are tested and selected for their ability to convey meaning to other individuals in the community~\cite{Steels2012_Experiments}. 
Individuals with cultural learning capabilities are able to observe and reproduce linguistic behaviours by understanding and internalising the intentional thinking behind their use (see Figure~\ref{fig:cultural_evo}; \citealp{Tomasello1999_Cultural}). 
This cultural evolutionary process is thought to be the main driver for the emergence of language in human populations~\cite{Tomasello2008_Origins}. 
Thus, the question we tackle in this paper is the following: \textit{Can we model the cultural evolution of linguistic conventions with neural agents?}

To this end, we propose a new method that enables a population of bi-directional agents to converge on a shared symbolic system for solving a reference game. Agents are trained to optimise mutual understanding in a fully decentralised way. 
We show that this model scales well to high-dimensional image datasets and large population sizes, and allows to study specific dynamics of cultural evolution. We show that the level of conventionality in a population is directly linked to how easy the language is to learn, thus favouring cultural transmission. Additionally, we demonstrate the value of learning from every individual interaction, instead of using batched interactions as is commonly done in the literature. 
Overall, this work provides a new way to model language emergence with neural agents, complementary to previous approaches, and should pave the way to more diverse experiments on neural emergent communication.





\section{Related Work}
\label{sec:RelatedWorks}

\paragraph{Computational models of language emergence.}
Since the 1990s, researchers in computational linguistics, cognitive science and artificial intelligence have employed agent-based models to simulate the emergence of languages~\cite{Steels1995_SelfOrgVoc, Steels1997_Synthetic, Vogt2005_EmergenceCompo}. This was in reaction to the unavailability of historical records of such language emergence in humans\,--\,for obvious reasons. 
The synthetic ethology approach allows us to model the origins of linguistic traits by observing the emerging communicative strategies in populations of artificial agents~\cite{Cangelosi2002_Simulation}. This enables the implementation and testing of theories on the emergence of particular natural language traits such as compositionality~\cite{Brighton2002_Compositional}, word order~\cite{Steels2015_Syntax, VanEecke2018}, case marking~\cite{Beuls2013_GrammaticalAgreement, Trijp2016_Case}, or speech~\cite{Boer2000_SelfOrga}. 

Central to this line of work is the idea that cultural evolution shapes the emergent languages. Because each individual in the population has its own knowledge of their previous communicative interactions, each new interaction becomes a transmission experience between two or more individuals. This has been extensively described in the iterated learning model~\cite{Smith2003_CompoEmerg, Smith2008_Cultural}, where this cultural transmission process is often portrayed with successive generations of agents learning a language taught by previous generations from limited examples~\cite{ScottPhillips2010_Lab}. This creates a transmission bottleneck that favours linguistic structures that are more learnable~\cite{Brighton2005_Learnability}. We aim to demonstrate this cultural transmission effect in populations of neural agents.

\paragraph{Cultural evolution in neural emergent communication.}
A large variety of neural models of language emergence have been proposed in the last decade. 
However, these models employ techniques that preclude capturing the dynamics of cultural evolution at play in language emergence~\cite{Galke2022_Missing}: using
agents that can only speak or listen~\cite{Chaabouni2022_Scale, Rita2022_PopHetero, Michel2023_Revisiting, Lee2024_OneToMany}, having populations of only two agents~\cite{Cao2018_Negotiation, Choi2018_CompComm, Cheng2023_Imit, Rita2022_GenOverf}, sharing gradients or internal representations between agents~\cite{Tucker2022_VQVIB, Gualdoni2024_Bridging}, or having a combination of these design choices. 
These choices prevent fully capturing dynamics of cultural evolution, because cultural evolution specifically works in populations of distributed individuals able to try, observe and reproduce communicative behaviours~\cite{Tomasello1999_Cultural}. 
Notable exceptions are \citet{Graesser2019} and \citet{Lian2024_NeLLComX} who both propose similar models to ours but use supervised learning on human language. 
Neural iterated learning models have also been proposed~\cite{Chaabouni2019_WordOrder, Ren2020_NIL}, but focus on vertical transmission between successive one-agent generations. 

Our work is the first to train populations of bi-directional agents in a fully distributed way and without human language supervision. 
This approach provides a new perspective on how cultural evolution drives the emergence of linguistic conventions in neural emergent communication. 






\section{Problem Definition}
\label{sec:Problem}

To assess how cultural evolution can lead to the emergence of linguistic conventions, we define our problem to explicitly consider the crucial elements that enable a cultural evolution process. Namely:

\paragraph{Populations.} 
Cultural evolution occurs in communities of individuals interacting towards common goals~\cite{Mesoudi2011_Cultural}. 
Each individual participates by trying out strategies and observing the strategies of others. This leads to the evolutionary process in which, at the population level, unsuccessful strategies are discarded in favour of successful ones. 
Different population structures may affect the dynamics of this cumulative co-adaptation process~\cite{Raviv2019_Communities}.

\paragraph{Situated local interactions.}
Communicative interactions take place locally between agents that observe the same scene\,--\,albeit from their own perspective~\cite{Beuls2024_Situated}. An interaction is characterised by the communicative need of 
agents to solve a particular task in their environment. 
By trying to solve this task, the agents test a strategy and evaluate it. Throughout many local interactions, agents accumulate experience and collectively converge towards conventions on how to communicate about entities of the world.

\paragraph{Distributed cognition.} 
To evaluate realistic accounts of cultural evolution, cognition must be fully distributed. Agents each have their own internal representations and they learn only from their observations\,--\,sensory inputs, exchanged messages and feedback. Agents cannot access the internal representations of other agents, and there is no central controller orchestrating how agents should behave or what conventions should look like. The only shared signals are the symbols used for communication and the feedback to the success of interactions~\cite{VanEecke2022_LangGameMARL}. While this is a hard constraint for learning in a multi-agent system, this realistically models the strong pressure on symbolic systems to carry complex meaning through discrete symbols.

\paragraph{Bi-directional communication capabilities.}
Language evolution takes place through the successive sharing of cultural artefacts~\cite{Tomasello1999_Cultural, Mufwene2019}. This requires individuals to be able to both produce these artifacts and understand them~\cite{Pickering2013_Integrated}. This interchangeability is a key property of human communication~\cite{Hockett1960_OriginsSpeech} and languages have been shown to evolve through this ability to observe and reproduce communicative strategies~\cite{Andersen1973}. 

In this paper, our goal is to implement these elements of cultural evolution in a neural model of language emergence, and demonstrate how they provide sufficient pressure for the emergence of efficient linguistic conventions.

\section{In-Context Reference Game}

Our problem definition is instantiated in an \textit{in-context reference game} where two agents are put in a shared scene and one agent\,--\,the speaker\,--\,needs to refer to an object in the scene by uttering a word to the other agent\,--\,the listener\,--, which then needs to point to the referred object. This setup has been used previously to highlight the role of contextual information in communication strategies~\cite{Steels1999_Situated, Botoko2024_Concept, Gualdoni2024_Bridging, Glowka2024_Contextdependent, Kobrock2024_Context, Zhang2025_NeLLComLex}. The meaning of linguistic utterances depends on the context in which they are uttered~\cite{Wittgenstein1953, Grice1975_Logic}. The in-context reference game allows us to model this important part of linguistic interactions. 

Concretely, we define a population of $N$ agents and a set of entities $W=\{e_1, ..., e_K\}$. An entity $e_i$ is an object represented by a feature vector of dimension $O$. The game proceeds as a sequence of interactions. An interaction starts by sampling two agents from the population and assigning them the roles of speaker and listener, respectively denoted as $S$ and $L$. The agents observe a scene $e=\{e_1,...,e_C\}\subset W$, with $C$ being the context size, i.e. the number of entities in the scene, and $e_i\in\mathbb{R}^O$. The topic of the interaction is always the last entity in the scene, i.e. $e_t=e_C$. The speaker observes the scene and produces a symbol $w_S$ to discriminate the topic $e_t$ from the other entities in the scene. The listener observes the scene, with entities randomly shuffled, and the uttered symbol $w_S$. Its goal is to infer which entity is the topic by adequately interpreting the speaker's message. If the listener's prediction is correct, agents are rewarded for a successful interaction. Otherwise, the interaction fails and the topic is disclosed to the listener. Based on this feedback, agents can update their strategies for future interactions. 

\begin{figure*}[t]
    \centering
    \includegraphics[width=\linewidth]{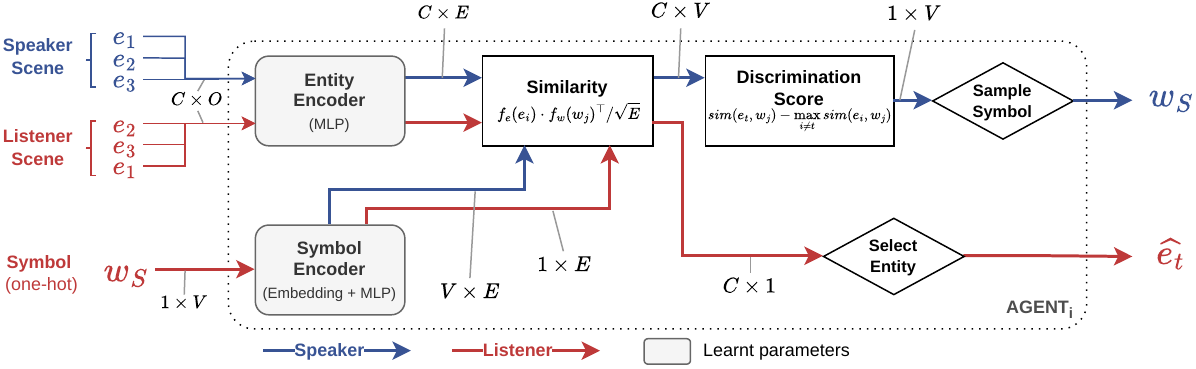}
    \caption{Architecture of an agent. The agent's Speaker and Listener forward passes are both represented by blue and red inputs and arrows. The learnt parameters are located in the two Encoders and are all used in both the Speaker and Listener roles. Dimensions of the matrices represented by arrows are shown. }
    \label{fig:archi}
\end{figure*}

\section{Method}
\label{sec:Method}

The goal of our methodology is to implement the necessary features for individuals in a population to enable the emergence of linguistic conventions through cultural evolution. 
That is, first, agents should be able to learn from every interaction and reuse the acquired knowledge in subsequent interactions, allowing a ratchet effect~\cite{Tennie2009_Ratcheting} where agents progressively bootstrap the linguistic convention. This is achieved through a neural network architecture designed to have a small number of parameters that are updated after each interaction (see Section~\ref{sec:Archi}). 
Second, agents should have cultural learning capabilities allowing them to maximise mutual understanding~\cite{Tomasello1999_Cultural, Tomasello2008_Origins}. This is implemented as an imitation objective, similar to previous work~\cite{Cheng2023_Imit}, where the listener agents explicitly learn to reproduce the intentional pattern observed during interactions (see Section~\ref{sec:Training}). 



\subsection{Agent Architecture}
\label{sec:Archi}

Agents are modelled as neural networks able to learn \textit{both} to produce symbols \textit{and} understand symbols coming from other agents. This bi-directionality is exploited in the architecture to learn representations that serve both the production and interpretation tasks. 
Figure~\ref{fig:archi} illustrates the architecture of an agent and how it is used for speaking and listening. 
Agents are equipped with two distinct neural networks: an \textbf{Entity Encoder} $f_e$, consisting of a two-layer multi-layer perceptron (MLP), to encode observed entities; and a \textbf{Symbol Encoder} $f_w$, encoding symbols with an embedding matrix followed by a two-layer MLP (see Appendix~\ref{app:archi} for more details). 
These two encoders represent all the learnt parameters of an agent, denoted $\theta_i$ for agent $i$. Their output embeddings are used to produce and comprehend symbols, as described below.

\paragraph{Speaker role.}
As a speaker, the goal is to utter a symbol that allows the listener to understand which entity is the topic in the current scene. To do so, it must assess how well fitted each symbol in its vocabulary is to discriminate the topic in the scene. It receives as input the scene $e\in\mathbb{R}^{C\times O}$ with the topic $e_t$ in the last position. Each entity in the scene $e_i$ is encoded \textit{separately} with the Entity Encoder, giving the entity embeddings $f_e(e|\theta_S)\in\mathbb{R}^{C\times E}$. Each symbol in the vocabulary is also encoded with the Symbol Encoder, giving the symbol embeddings $f_w(w|\theta_S)\in\mathbb{R}^{V\times E}$. From these two matrices, we compute the \textbf{similarity} between each entity and each symbol with a scaled dot-product:
\begin{equation}
    \Sim(e,w|\theta_{S})=\frac{f_e(e|\theta_{S})\cdot f_w(w|\theta_{S})^\top}{\sqrt{E}},
    \label{eq:sim}
\end{equation}
with $E$ the dimension of the embeddings. Scaling by $1/\sqrt{E}$ stabilises the magnitudes of the similarity values~\cite{Vaswani2017_Transformer}. 

Equation~\ref{eq:sim} gives the matrix $\Sim(e,w|\theta_S)\in\mathbb{R}^{C\times V}$ containing all the similarities between each pair of entity and symbol. As in \citet{Botoko2024_Concept}, we compute the \textbf{discriminative power} of each symbol $w_i$ for the topic $e_t$, as the difference between the similarity of $(e_t, w_i)$ and the maximum similarity between $w_i$ and the other entities in the scene, i.e.:
\begin{multline}
    \delta(e_t,w_i|\theta_S)=\Sim(e_t,w_i|\theta_S) \\ -\max_{j\neq t}\Sim(e_j,w_i|\theta_S).
    \label{eq:DiscrScore}
\end{multline}
This discrimination score measures how much each symbol is ``more similar to the topic than to the other entities''. 
To make the discrimination score fully differentiable with regard to all entities in the scene, we replace the $\max$ operation by the continuous approximation $LogSumExp$ ($LSE$):
\begin{align}
\begin{split}
    \delta(e_t,w_i|\theta_S) {}&= \Sim(e_t,w_i|\theta_S) \\ 
                    & \quad- LSE_{j\neq t}\bigl(\Sim(e_j,w_i|\theta_S)\bigr)
\end{split} \\
\begin{split}
    {} &= \Sim(e_t,w_i|\theta_S) \\
    & \quad- \log\sum_{j\neq t}e^{\Sim(e_j,w_i|\theta_S)/\tau_{lse}},
\end{split}
\end{align}
with $\tau_{lse}$ a temperature parameter that can be used to make $LogSumExp$ closer to the $\max$ function (in practice we keep it at 1). Using the $LogSumExp$ allows us to consider all the entities in the scene to compute the discrimination score and provides richer feedback during learning updates. 

The speaker's policy is the distribution over words in the vocabulary derived from the discrimination scores of all the words in its vocabulary:
\begin{align}
    \pi(w_j|e,\theta_S) &= softmax\bigl(\delta(e_t, w|\theta_S) / \tau_{s}\bigr)_j \label{eq:SpeakPolicy},
\end{align}
with $\tau_{s}$ a temperature parameter controlling the flatness of this distribution, which can be used to control the level of exploration in the speaker's policy. The speaker's output symbol $w_S$ is sampled from this distribution and sent as a one-hot vector to the listener.

\paragraph{Listener role.}
As a listener, agent $L$ has to interpret the received symbol and point to the correct entity in the scene. As for the speaker, it encodes the scene $e$ with its Entity Encoder: $f_e(e|\theta_L)\in\mathbb{R}^{C\times E}$. The received symbol $w_S$ goes through the Symbol Encoder to get $f_w(w_S|\theta_L)\in\mathbb{R}^{1\times E}$. From these embeddings, the similarity $\Sim(w_S, e_i|\theta_L)$ between the received word and each entity in the scene is computed as defined for the speaker. Based on these similarity values, the listener computes a probability distribution over the entities:
\begin{equation}
    p_L(e_i|e,w_S,\theta_L)=softmax\bigl(\Sim(e, w_S|\theta_L)\bigr)_i.
    \label{eq:ListenerPred}
\end{equation}
The listener then selects the entity with the highest similarity to $w_S$ as its prediction $\widehat{e_t}$. If $\widehat{e_t}=e_t$, the listener has pointed to the right entity and the interaction is successful. Otherwise, the interaction is a failure. 

\subsection{Training}
\label{sec:Training}

After each interaction, agents are trained to improve their success in upcoming interactions. The two participating agents receive a reward $R=1$ for success or $R=0$ for failure. In case of failure, the topic entity is also disclosed to the listener, allowing to compute a supervised learning objective. 
Based on this feedback and the events of the interaction, the speaker and listener agents compute their learning objective separately and update their parameters accordingly. 

\paragraph{Speaker objective.} 
The speaker's goal is to choose the words that will maximise the success of its interactions. This is defined as the REINFORCE objective~\cite{Williams1992_Reinforce}:
\begin{equation}
    \mathcal{L}_{speak}(\theta_S)=-\log\bigl(\pi(w_S|e,e_t,\theta_S)\bigr)R.
    \label{eq:LossSpeak}
\end{equation}
Minimising this objective will increase the probabilities assigned by $\pi(.|\theta_S)$ to successful symbols. The full derivation of this objective is provided and analysed in Appendix~\ref{app:DiscrScore}.

In failed interactions, $R=0$, so there is no update for the speaker. This shows an intended bias of this approach towards the speaking agent's choice of symbols to utter. The choices of speaker agents are considered valid, while not necessarily perfect. 

\paragraph{Listener objective.}
The listener's primary goal is to interpret the received symbol correctly and point to the right entity. This is optimised by minimising the cross-entropy loss:
\begin{equation}
    \mathcal{L}_{pred}(\theta_L) = -\log\bigl(p_L(e_t|e,w_S,\theta_L)\bigr).
    \label{eq:LossPred}
\end{equation}
Minimising $\mathcal{L}_{pred}$ will increase the similarity of the listener's representations of $e_t$ and $w_S$.

In addition, the listener also tries to imitate the speaker's communication strategy
by minimising the cross-entropy loss on its speaking policy $\pi(.|\theta_L)$, with the target output being the speaker's symbol:
\begin{equation}
    \mathcal{L}_{imit}(\theta_L) = -\log\pi(w_S|e,e_t,\theta_L).
    \label{eq:LossImit}
\end{equation}
Because the speaking policy $\pi(.|\theta_L)$ uses the discrimination score (see Equation~\ref{eq:SpeakPolicy}), the imitation objective provides a rich learning signal for the listener to update its representations. 

The complete objective for the listener is therefore:
\begin{equation}
    \mathcal{L}_{L}(\theta_L)=\beta_{p}\mathcal{L}_{pred}(\theta_L)+\beta_i\mathcal{L}_{imit}(\theta_L),
\end{equation}
with $\beta_p$ and $\beta_i$ being two coefficient hyperparameters controlling the weights of each objective.
\section{Experimental Setup}
\label{sec:XPsetup}

\paragraph{Datasets.} 
We instantiate the in-context reference game in four different domains: the \textit{ColorReference} dataset~\cite{Monroe2017_Pragmatic}, with 47K scenes of three colors; the \textit{ManyNames} dataset~\cite{Silberer2020_ManyNames}, as processed by \citet{Gualdoni2024_Bridging}, with 25K naturalistic scenes composed of an image with one topic object and one distractor (see the details of data extraction in the original paper); \textit{CelebA}~\cite{Liu2015_CelebA}, with 202K images of celebrity faces; and \textit{MSCOCO}~\cite{Lin2014_MSCOCO}, with 159K images of everyday scenes. 
For \textit{CelebA} and \textit{MSCOCO}, images are first encoded by a pre-trained DINOv2~\cite{Oquab2024_DINOv2} model and scenes are constructed by randomly sampling $C$ images and setting one as the topic of the interaction. 
Details about pre-processing and splits are available in Appendix~\ref{app:data}.

\paragraph{Training details.} 
Agents are updated after each interaction they take part in. Training is fully decentralised. Unless specified, interactions are not batched, meaning that agents use only one interaction to compute their objectives. 
Our model uses $\beta_p=1$ and $\beta_i=0.2$, which were found to give the best trade-off between success and conventionality. 
Implementation details and other hyperparameters are provided in Appendix~\ref{app:train_details}. 
Our code is available at: \href{https://gitlab.ai.vub.ac.be/mtoquebiau1/cultural-evo}{gitlab.ai.vub.ac.be/mtoquebiau1/cultural-evo}.

\paragraph{Evaluation.} 
During evaluation, we evaluate the population by randomly sampling 1000 scenes from the held out test sets. All agents in the population are evaluated as speakers on these scenes, with randomly-selected listeners. 
We measure the average \textbf{communicative success} to track the efficiency of solving the game. To assess whether the evaluated populations converge to shared communicative strategies, we measure the level of \textbf{conventionality} of the emerged languages. We define the \textit{population coherence} metric that looks at which symbols were used by the agents and computes the probability of having two agents utter the same symbol when observing the same scene (see Appendix~\ref{app:conv} for detailed computation). Note that other metrics for conventionality have been proposed in the past for multi-symbol utterances~\cite{Raviv2019_Communities, Rita2022_PopHetero, Michel2023_Revisiting, Kim2021_Varying}. Our population coherence measure is equivalent to them for single-symbol utterances. 
Finally, we track the \textbf{number of distinct symbols} that account for $95\%$ of all utterances, denoted $V_{used}^{95\%}$, to evaluate the size of the effective vocabulary in the emerged languages. 

\section{Experiments}
\label{sec:XP}

\begin{table}[t]
    \small
    \centering
    \resizebox{\linewidth}{!}{%
    \begin{tabular}{ccccc}
    \toprule
    \vspace{0.02em}
    \textbf{Method}  & \textbf{Success} (\%)  & \textbf{Conv.} (\%) & $V_{used}^{95\%}$ \\
    \midrule
    \midrule
    \multicolumn{4}{l}{\textbf{\textit{ColorReference} ($C=3$, $O=3$)}} \\
    \midrule
    Ours            & 99.13$_{\pm 0.15}$             & \textbf{93.18$_{\pm 0.59}$}  & 8.8$_{\pm 0.42}$   \\
    No-Imitation    & \textbf{99.54$_{\pm 0.05}$}    & 76.46$_{\pm 1.72}$           & 20.5$_{\pm 1.51}$  \\
    Similarity      & 85.46$_{\pm 0.78}$             & 52.4$_{\pm 9.17}$            & 21.9$_{\pm 4.25}$  \\
    S-L             & 97.36$_{\pm 0.35}$             & 5.84$_{\pm 2.6}$             & 30.4$_{\pm 2.41}$  \\
    
    \midrule
    \multicolumn{4}{l}{\textbf{\textit{ManyNames} ($C=2$, $O=512$)}} \\
    \midrule
    Ours            & \textbf{99.61$_{\pm 0.13}$}   & \textbf{94.11$_{\pm 0.37}$}  & 4.1$_{\pm 0.32}$   \\
    No-Imitation    & 99.11$_{\pm 0.15}$            & 53.81$_{\pm 4.42}$           & 13.6$_{\pm 1.07}$  \\
    Similarity      & 82.17$_{\pm 11.74}$           & 15.77$_{\pm 8.14}$           & 19.5$_{\pm 3.95}$  \\
    S-L             & 97.91$_{\pm 0.2}$             & 1.55$_{\pm 1.24}$            & 19.9$_{\pm 1.2}$   \\
    
    \midrule
    \multicolumn{4}{l}{\textbf{\textit{CelebA} ($C=10$, $O=384$)}} \\
    \midrule
    Ours            & 97.41$_{\pm 0.66}$            & \textbf{82.74$_{\pm 1.17}$}   & 35.9$_{\pm 2.51}$  \\
    No-Imitation    & \textbf{98.63$_{\pm 0.28}$}   & 68.42$_{\pm 1.8}$             & 76.57$_{\pm 5.68}$ \\
    Similarity      & 38.07$_{\pm 25.09}$           & 15.51$_{\pm 13.06}$           & 29.2$_{\pm 17.61}$  \\
    S-L             & 88.54$_{\pm 1.46}$            & 13.83$_{\pm 2.7}$             & 62.3$_{\pm 4.57}$  \\
    
    
    \midrule
    \multicolumn{4}{l}{\textbf{\textit{MSCOCO} ($C=10$, $O=384$)}} \\
    \midrule
    Ours            & 96.53$_{\pm 0.5}$             & \textbf{81.91$_{\pm 1.59}$}   & 32.5$_{\pm 1.35}$  \\
    No-Imitation    & \textbf{97.91$_{\pm 0.47}$}   & 66.51$_{\pm 1.64}$            & 77.6$_{\pm 1.35}$  \\
    Similarity      & 9.73$_{\pm 0.24}$             & 1.3$_{\pm 1.53}$              & 9.7$_{\pm 1.16}$   \\
    S-L             & 82.55$_{\pm 0.85}$            & 10.05$_{\pm 1.28}$            & 57.5$_{\pm 2.32}$  \\
    \toprule
    \end{tabular}%
    }
    \caption{Results of our model and baselines on the evaluation sets of each dataset, after 2M training interactions, with 10 agents (10 runs each, mean and standard deviation shown). Context size $C$ and entity dimension $O$ are shown between parentheses.}
    \label{tab:baselines}
    \vspace{-1em}
\end{table}

\subsection{Model performance}
\label{sec:baselines}

We start by evaluating how our model performs in the four datasets and how it compares against two ablations: \textbf{No-Imitation} which does not use the imitation loss in the listener's objective, and \textbf{Similarity} which uses the similarity scores (Eq.~\ref{eq:sim}) instead of the discrimination scores to compute the speakers' policy. In the Similarity baseline, speaker agents select their utterance based only on how similar symbols in their vocabulary are to the topic, thus not considering the other entities in the scene. 

Table~\ref{tab:baselines} shows the results after 2M interactions for populations of 10 agents. Our approach reaches near-perfect success on all datasets. 
The No-Imitation baseline is also highly successful with the highest score in three out of four settings. However, the conventionality of the emerged languages is significantly higher for our model than without imitation. This shows that an explicit imitation mechanism helps neural agents converge towards common ground, as it is thought to occur in humans~\cite{Tomasello1999_Cultural}. 
The Similarity baseline performs particularly poorly in settings with a large context size $C$, showing that the in-context reference game benefits from the use of the discrimination score to efficiently learn context-based representations. It still obtains acceptable success rates on \textit{ColorReference} and \textit{ManyNames}, where $C$ is small, but with significantly less conventionality than in our model. These results show that the discrimination score helps agents converge on similar representations by allowing richer learning updates. 

\subsection{Bi-directional vs. Speaker-Listener}
\label{sec:Speaker-Listener}

Next, we demonstrate the importance of having bi-directional agents by introducing a \textbf{Speaker-Listener} (S-L) baseline where  half of the population always act as speakers and the other half as listeners\footnote{The S-L population being $N/2$ speakers and $N/2$ listeners ensures that S-L agents have the same number of training updates as in our model. In our model, agents have a $2/N$ chance of being selected for an interaction (either as speaker or listener), and in S-L, speakers (resp. listeners) have a $\frac{1}{N/2}=2/N$ chance of being selected.}. As shown in Table~\ref{tab:baselines}, the S-L baseline obtains high success on all datasets, though significantly lower than our model, but with low levels of conventionality. This indicates that S-L populations are able to solve the in-context reference game without converging to a single shared language. 

\begin{figure}[t]
    \centering
    \includegraphics[width=\linewidth]{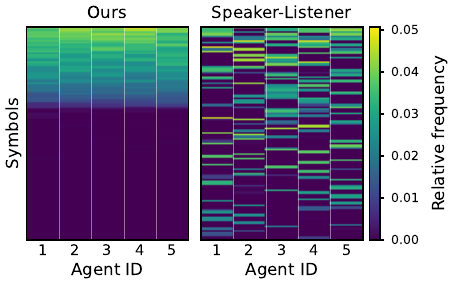}
    \caption{Relative usage frequency of symbols (rows) across 5 agents (columns) in the \textit{CelebA} setup. Our model produces highly aligned symbol usage across agents, whereas the S-L baseline exhibits speaker-specific usage patterns, despite both populations reaching 95\% success.}
    \label{fig:usage_top_100}
    \vspace{-1em}
\end{figure}

To further visualise this, we compare the relative usage frequency of symbols across agents in our model and the S-L baseline, as shown in Figure~\ref{fig:usage_top_100}. Both variants are trained on \textit{CelebA} with 10 agents until reaching 95\% success, requiring approximately 500K interactions for our model and 20M for S-L (see Appendix~\ref{app:usage_details} for details on the construction of this figure). Despite achieving similar communicative success, conventionality is substantially higher in our model (74\% vs. 16\%). Agents in our model converge towards highly similar usage distributions, whereas S-L speakers rely on different subsets of symbols. This highlights how, in S-L populations, the listeners learn to adapt to the individual language of each speaker. On the other hand, with bi-directional agents, the constant transmission of communicative behaviours leads to a single well-conventionalised language.

Note that the S-L baseline does not use the imitation objective as, conceptually, it does not make sense to train listeners to play the role of speaker. However, to ensure a fair comparison, we also provide results of the S-L baseline trained with the imitation objective on CelebA, with the same $\beta_i=0.2$ as in our model, with 10 independent runs (see Appendix~\ref{app:SL+imitation} for further results with various values of $\beta_i$). Compared to our model, imitation has a similar but weaker effect on S-L. Success decreases slightly, from $88.54\%$ to $86.14\%$. Conventionality is improved from $13.83\%$ to $18.94\%$. This gain is smaller than the one observed from No-Imitation to our model (from $68.42\%$ to $82.74\%$). 

Overall, these results clearly indicate that S-L models do not produce significant pressure toward conventionality (see Appendix~\ref{app:vocab_size} for results when varying $V$). S-L listeners are able to adapt to each speaker, even with a relatively small neural network architecture, and an imitation objective does not solve this issue. We show that having agents alternate speaking and listening roles, and learning representations that serve both of these roles, creates sufficient pressure toward conventionality. As listeners will act as speakers in future interactions, reproducing previously observed behaviours is now advantageous and successful behaviours will be selected, as in a typical cultural evolution process.


\subsection{Scaling Populations}

We investigate how well our model scales with the population size. Previous work has experimented with populations of up to 100 agents~\cite{Chaabouni2022_Scale}. Thus, we compare populations of 10, 100 and 1000 agents on \textit{ManyNames}.
For a fair comparison, we train populations until they reach an average of 100K interactions per agent: 500K interactions for 10 agents, 5M for 100, and 50M for 1000 (see Appendix~\ref{app:pop} for more details and full results).
Populations of 10, 100 and 1000 agents respectively obtain $98.97\%$, $97.4\%$ and $95.89\%$ average success and $91.23\%$, $91.62\%$ and $92.01\%$ average conventionality. 
The slight loss in success in larger populations can be understood as agents encounter more interlocutors and knowledge takes more time to be shared in the whole population. 
Nevertheless, this strong performance across all population sizes indicates that the model scales effectively.
\subsection{Conventionality and Learnability}


\begin{figure}[t]
    \centering
    \includegraphics[width=\linewidth]{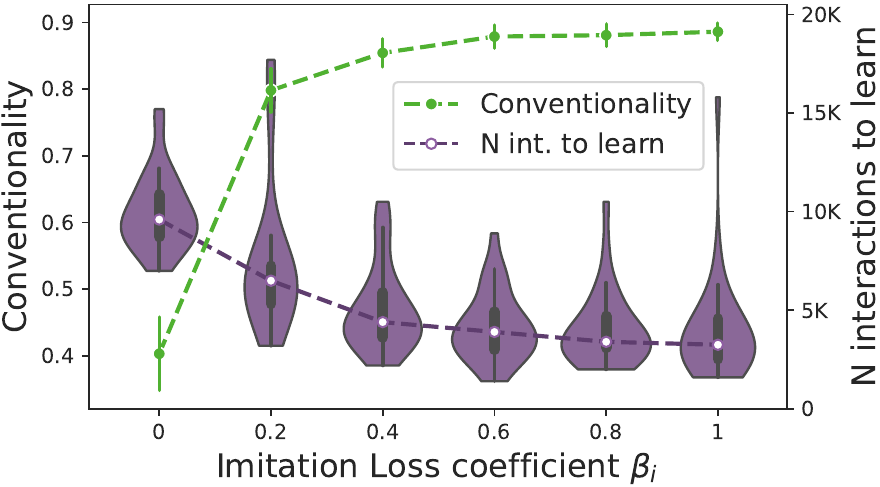}
    \caption{Learnability experiment in \textit{ColorReference} with 100 agents, for different values of $\beta_i$ (30 independent runs for each $\beta_i$). Green line shows mean conventionality when populations reach 95\% success, with standard deviation in error bars. Violin plots show the distribution of the time to learn the language, with the dashed purple line indicating the median.}
    \label{fig:conv}
    \vspace{-1em}
\end{figure}

To highlight the role of conventionality in the cultural evolution of a language, we conduct a learnability experiment~\cite{Li2019_Ease, Wright2026_Conventionality} to show that more conventional languages are easier to learn. 
We train populations of agents with different values of the imitation loss coefficient $\beta_i$, until they reach 95\% success. Then, a new agent is added to the population and trained to understand the emerged language by playing as a listener (in all cases, the new agent is trained with $\beta_i=1$). In Figure~\ref{fig:conv}, we plot the conventionality of the emerged languages and the number of interactions for the new agents to reach 95\% success, with 30 independent runs for each value of $\beta_i$.

These results first show that conventionality increases with $\beta_i$. We also observed a small loss in success when increasing $\beta_i$. This confirms a trade-off between conventionality and success, as already observed in the comparison with the No-Imitation baseline (see Table~\ref{tab:baselines}). Second, more conventional languages are easier to learn, going from 10K interactions to learn on average for $\beta_i=0$ to less than 5K for $\beta_i=1.0$. 
In conventionalised languages, the learner agent has fewer symbols to learn (see $V_{used}^{95\%}$ in Table~\ref{tab:baselines}) and observes them in action more consistently. 
This clearly highlights the reciprocal link between conventionality and transmission in cultural evolution, with each favouring the other.  

 \subsection{Batched Interactions}
 \label{sec:batch}

\begin{figure}[t]
    \centering
    \includegraphics[width=\linewidth]{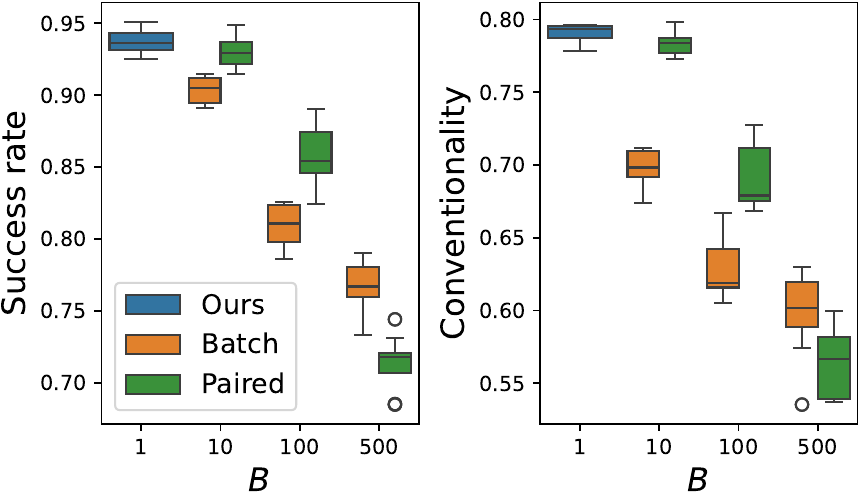}
    \caption{Success and conventionality after training on \textit{CelebA} for 10M interactions, with 100 agents, comparing our model ($B=
    1$) to ``batched'' and ``paired'' variants (10 runs each).}
    \label{fig:batch}
    \vspace{-1em}
\end{figure}

Unlike previous work, our approach does not use batched interactions. Usually, agents interact over batches of scenes, with batch size $B$ ranging from 32 to 1024, meaning that the two agents observe $B$ different scenes, interact on all scenes independently, and finally compute an averaged optimisation objective given the experience gathered in the $B$ interactions. Learning from a batch of samples is a common practice in deep learning to improve the stability of the learning updates. In our model, we do not use batched interactions to allow agents to learn from every interaction and immediately reuse the acquired knowledge in subsequent ones. 
To evaluate how this affects training, we compare our model ($B=1$) against baselines using $B=\{10, 100, 500\}$. Since the learning rate may depend on batch size, we tuned it separately for each value of $B$, selecting the learning rate that achieves the best trade-off between success and conventionality (see Appendix~\ref{app:batch} for details).

Figure~\ref{fig:batch} shows the results on \textit{CelebA}, with 100 agents, after training for 10M interactions. Training with batched interactions clearly leads to lower success and conventionality, with both decreasing as $B$ increases. We hypothesise that the batched versions suffer from two problems: agents are updated less often because interacting agents only update once after $B$ interactions, and agents are paired for several interactions which may lead to more specialisation to one interlocutor. To investigate these issues, we experiment with an ablation where agents are also paired for $B>1$ interactions, but are updated after each interaction with the same learning rate as our model. As shown in Figure~\ref{fig:batch}, the success of ``paired'' baselines also decreases when increasing $B$. Interestingly, below $B=500$, the paired baselines are better than the batched ones, showing that agents benefit from training after each interaction, even when they are paired. 
Overall, this highlights how changing partners and updating as often as possible are important to drive the cumulative bootstrapping of linguistic conventions.

\subsection{Symbol Usage Analysis}
\label{sec:word_usage}

\begin{figure}[t]
    \centering
    \includegraphics[width=\linewidth]{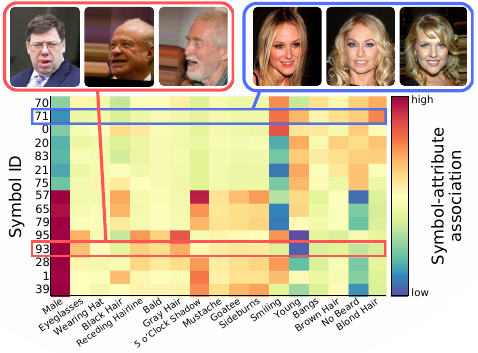}
    \caption{Co-occurrences between symbols and \textit{CelebA} attributes for the 15 most frequently used symbols, computed from 100K interactions on unseen images (see Appendix~\ref{app:word_usage} for an extended version). Warmer colours indicate stronger positive associations between symbols and attributes. Symbols 71 and 93 are illustrated with images that most strongly elicited these symbols.}
    \label{fig:word_usage}
    \vspace{-1em}
\end{figure}

To analyse the semantic structure of the emerged conventions, we evaluate trained agents on 100K interactions on unseen \textit{CelebA} images and record the symbols produced. Although agents only observe the images, the dataset provides human-annotated visual attributes for each image (e.g. \textit{wearing hat}, \textit{smiling}), which we then use to plot the co-occurrences of these visual attributes and the symbols used by the agents (see Figure~\ref{fig:word_usage}). We find that symbols capture bundles of correlated attributes, rather than isolated concepts. For example, symbol 71 is associated with \textit{smiling}, \textit{young} and \textit{blond hair}, among others, whereas symbol 93 is associated with \textit{male}, \textit{gray hair} and not-\textit{young}. These associations are also substantially more consistent across agents than those in the S-L baseline.
\section{Discussion and Conclusion}
\label{sec:Discussion}


We presented a neural approach to language emergence aimed at modelling the dynamics of cultural evolution through local situated interactions among agents in a population. This is achieved by modelling agents with bi-directional communication capabilities, where representations are shared for speaking and listening, and where agents are trained in a distributed fashion with an imitation objective inspired by cultural learning in humans. Under these conditions, agents are able to solve an in-context reference game in high-dimensional domains and across large populations.

The proposed model stands out for its sample efficiency, allowing agents to learn from every interaction and reuse acquired knowledge immediately after each interaction. This enables the ratcheting effect characteristic of cultural evolution, with agents adapting to each other and successful behaviours being selected at the population level. We showed that this does not happen in speaker-listener populations, where, instead, listeners adapt to the individual languages of every speaker agent. 
This clearly indicates that neural agents have the learning capacity to solve language games without cultural evolution. However, under the right conditions, cultural evolution does shape the emergent languages in populations of neural agents, by favouring conventionalised communicative behaviours. Ultimately, the emerged languages benefit from these conventions by having more compact, effective vocabularies and being easier to learn for new agents.

This work shows the importance of investigating the cultural pressures at play in language emergence, and provides methodological guidelines to successfully implement these pressures in neural models of emergent communication. In this respect, our approach supports recent efforts advocating for greater attention to the modelling assumptions and conditions under which languages emerge and evolve~\cite{Beuls2024_Situated, Galke2025_Lessons}.
The proposed method is, therefore, complementary to previous work focused on cognitive biases in language use and calls for new work investigating the interplay between the cultural and cognitive pressures for language emergence.

\section*{Limitations} 

While this work explores the emergence of human-like languages through cultural evolution, it does not paint the full picture. 
One missing element is the ability to express meaning with multi-word utterances. Agents in our model only utter a single symbol. This is sufficient to study how conventions emerge for associating meaning from the world with symbolic representations.
A natural extension of our model would modify the agent speaking policy with a recurrent neural network to allow the generation of sequences of symbols as utterances. 
This would allow us to model the emergence of compositional structures in utterances~\cite{Chaabouni2020_Composit, Galke2024_Composit}. 

For the sake of simplicity, we chose to focus on paired interactions. However, human communication can occur with more than two interlocutors, as explored previously with speaker-listener populations~\cite{Lee2024_OneToMany}. Such grouped interactions may play an interesting role in the cultural evolution of language. We hypothesise that interactions involving one speaker and $k$ listeners would improve the convergence speed and the conventionality level. 

Finally, a limitation of our model is the fact that we need to set a fixed vocabulary size. This is common practice in neural emergent communication~\cite{Rita2025_LangEvoDeep}, because neural networks are usually initialised randomly at the start of each experiment. This introduces a bias that may have important effects on the results, as the size of the vocabulary determines the search space for the speaker policy. In this work, we chose to use $V=100$ for all experiments, but other settings might require tuning $V$. This also limits our ability to study the role of invention of new symbolic forms to solve difficult communicative interactions~\cite{Tomasello1999_Cultural, Skyrms2010}. Here, invention occurs ``by default'' at the start by initialising random parameters for the Symbol Encoder of each agent. An interesting avenue would be to adapt the neural network architecture to allow new vocabulary entries to be added during the course of the game.




\section*{Acknowledgements}

We are grateful to Liesbet De Vos, Lara Verheyen and Jamie D. Wright for their insightful feedback on the manuscript, and to the anonymous ARR reviewers, meta-reviewer and area chair for their constructive comments and suggestions. This research was supported by the F.R.S.-FNRS-FWO WEAVE project HERMES I under grant numbers T002724F (F.R.S.-FNRS) and G0AGU24N (FWO), the Flemish Government
under the Onderzoeksprogramma Artificiële Intelligentie (AI) Vlaanderen programme and the AI
Flagship project ARIAC by DigitalWallonia4.ai.




\bibliography{refs}

\appendix

\begin{table*}[t]
\small
    \centering
    \setlength{\tabcolsep}{6pt}
    \renewcommand{\arraystretch}{0.95}
    \begin{tabular}{ccccc}
        \toprule
        \multirow{2}{*}{\textbf{Hyperparameter}}  & \multicolumn{4}{c}{\textbf{Method}} \\ 
        \cmidrule(lr){2-5} 
        & Ours     & No-Imitation     & Speaker-Listener  & Similarity \\ \midrule
        Learning rate $\alpha$  & \multicolumn{4}{c}{0.0001}  \\
        Vocabulary size $V$  & \multicolumn{4}{c}{100}  \\
        Hidden dimension $H$  & \multicolumn{4}{c}{64}  \\
        Embedding dimension $E$  & \multicolumn{4}{c}{64}  \\
        \textit{LogSumExp} temperature $\tau_{lse}$  & 1.0 & 1.0 & 1.0 & -  \\
        Speaker policy temperature $\tau_{s}$  & \multicolumn{4}{c}{0.1}   \\
        Prediction loss coefficient $\beta_{p}$  & \multicolumn{4}{c}{1.0}   \\
        Imitation loss coefficient $\beta_{i}$  & 0.2  & 0.0  & 0.0 & 0.0    \\ 
        Entropy max coefficient  & -  & -  & - & 0.1    \\ 
        \bottomrule 
    \end{tabular}
    \caption{Hyperparameters used in all experiments, unless specified otherwise.}
    \label{tab:hyperparam}
\end{table*}

\section{Implementation details}
\label{app:imp_details}

\subsection{Agent Architecture}
\label{app:archi}


The \textbf{Entity Encoder} and \textbf{Symbol Encoder} are the agent's neural networks. The Entity Encoder encodes each entity in a scene separately to an embedding with a multi-layer perceptron (MLP): $f_e:\mathbb{R}^O\rightarrow\mathbb{R}^E$. The Symbol Encoder encodes one-hot symbols with an embedding matrix followed by an MLP: $f_w:\mathbb{R}^V\rightarrow\mathbb{R}^E$. 

The two MLPs in the encoders consist of two linear layers, each followed by a ReLU activation and a Layer Normalisation. They both use a hidden dimensionality $H$ of 64. The embedding matrix in the Symbol Encoder maps one-hot vectors to embeddings of the same dimensionality as the hidden layers: $H=64$. 

This results in a rather lightweight architecture. For reference, a population of 10 agents playing on the \textit{ManyNames} dataset ($O=512$) with $V=100$ comprises 522,240 learnable parameters. 



\subsection{Training Details}
\label{app:train_details}

Agents are trained after each interaction they take part in. To update the parameters, we use an RMSProp optimiser. The gradients of the speaker are always clipped so that their norm is no higher than 10. We found that it consistently led to better results. We hypothesise that this is because the update from REINFORCE is too noisy. All the hyperparameters we use are provided in Table~\ref{tab:hyperparam}. 

As shown in the table, the Similarity baseline does not use the imitation objective. This is because we observed that imitation consistently led to the collapse of the Similarity agents' speaking policies to using only one symbol, thus leading to chance-level success. Unlike the main model and other baselines, it also uses an entropy maximisation objective~\cite{Williams1991_MaxENT} for the speaking policy to incite the exploration of different symbols. We found that it was required for the Similarity baseline to reliably converge to acceptable success rates. Note that the use of such an entropy maximisation objective is common in emergent communication~\cite{Chaabouni2022_Scale, Michel2023_Revisiting, Zhang2025_NeLLComLex}. The fact that our model does not need it to solve the task indicates that the cultural evolutionary processes at play make policy exploration unnecessary. 

Experiments were run on a CPU cluster equipped with \textit{AMD EPYC 9535} processors. We estimate the total run time of our experiments to be around 480 hours.



\section{Datasets}
\label{app:data}


The \textit{ColorReference} dataset, consisting of human colour reference games over RGB colour patches, is used as provided by \citet{Monroe2017_Pragmatic}, without further pre-processing. For \textit{ManyNames}~\cite{Silberer2020_ManyNames}, a dataset of approximately 25K real-world images of everyday scenes, we follow the guidelines given by \citet{Gualdoni2024_Bridging} to download the dataset and extract visual features with a pre-trained ResNet18 encoder (see their \href{https://github.com/InfoCogLab/info-sem-prag-neurips2024}{code}), yielding feature vectors of dimension 512. All feature vectors are normalised between 0 and 1. For \textit{CelebA}~\cite{Liu2015_CelebA} and \textit{MSCOCO}~\cite{Lin2014_MSCOCO}, we use the original train/test splits of the datasets. \textit{CelebA} contains over 202K images of celebrity faces and \textit{MSCOCO} (v2017) contains over 159K naturalistic images of everyday scenes. We extract visual features with a pre-trained DINOv2~\cite{Oquab2024_DINOv2} model, yielding feature vectors of dimension 384. All feature vectors are normalised between 0 and 1. The pre-encoded and split datasets are available with the code online. All the datasets and models used are licensed under CC BY 4.0, except for DINOv2 which is licensed under Apache License 2.0, and CelebA which is "available for non-commercial research purposes only" as mentioned on their website.
\section{Conventionality Metric: Population Coherence}
\label{app:conv}

\begin{figure*}[t]
    \centering
    \includegraphics[width=\textwidth]{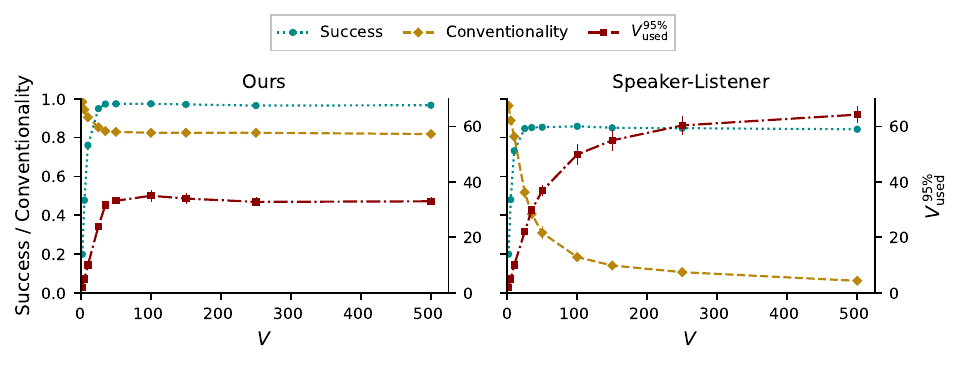}
    \caption{Effect of vocabulary size $V$. Small vocabularies ($V<35$) under-capacitate both models, limiting communicative success. Once sufficient capacity is available, our model converges to a compact active vocabulary. For all $V \geq 35$, $\sim 33$ active symbols emerge, maintaining high conventionality. In contrast, the Speaker-Listener baseline gradually expands its effective vocabulary as $V$ grows, while conventionality sharply decreases.}
    \label{fig:vocab-comparison}
\end{figure*}

Conventionality in a given population can be measured by how coherent agents in the population would be when faced with the same scene and topic. Previously, this was estimated with the ``linguistic coherence'' or ``pairwise coherence'', in which agents in an interaction would be considered ``coherent'' if they would utter the same word to describe the topic. 

This pairwise coherence can be extended to measure coherence at the population level. We define \textbf{population coherence} as the \textit{probability of having two randomly-sampled agents utter the same word when given the same scene and topic}. This can be computed for any given scene $e=\{e_1,...,e_C\}$, topic $e_t\in e$, and population $P$ of $N$ agents, by evaluating all agents in the population on the scene $e$. All words uttered $w$ by the agents are stored and their usage $u_w$ is computed as the number of times they were uttered by an agent. From this, we can compute population coherence with the following definition:
\begin{multline*}
    PC(e,e_t,P)=\sum_{w\in U}\bigl(p(\text{first agent utters }w)\\
    \times p(\text{second agent utters }w)\bigr),
\end{multline*}
with $U$ the set of uttered words. 
The two probabilities are computed using the recorded usage of each word:
\begin{align*}
    PC(e,e_t,P) &= \sum_{w\in U}\frac{u_w}{N}\times\frac{u_w-1}{N-1}\\
    &= \frac{1}{N(N-1)}\sum_{w\in U}u_w(u_w-1).
\end{align*}
This can be computed for a large set of scenes to measure the level of conventionality in the population. 

\section{Details on Experiments \& Additional Results}
\label{app:add_results}

\subsection{Varying the vocabulary size}
\label{app:vocab_size}


Previous work has shown that constraints on the communication channel can shape emergent languages.
In particular, discrete communication can favour lower-complexity systems \cite{chaabouni2021communicating} and vocabulary size can affect communicative success and learnability \cite{kunkele2025learning}. This suggests a trade-off: restricting the available vocabulary may increase pressure towards a shared convention by forcing agents to reuse a smaller set of symbols, but an overly small vocabulary may not provide enough capacity to solve the task.

We therefore vary the vocabulary size $V$ for both our model and the Speaker-Listener (S-L) baseline. In both settings, we train a population of 10 agents on \textit{CelebA} with $V \in \{2, 5, 10, 35, 50, 100, 150, 250, 500\}$ for two million interactions. Figure~\ref{fig:vocab-comparison} shows how the vocabulary size $V$ affects communicative success, conventionality and the effective vocabulary size ($V^{\,95\%}_{\mathrm{used}}$) in our model and the Speaker-Listener baseline.

In our bi-directional model, small vocabularies ($V \in \{2, 5, 10\}$) under-capacitate the agents, preventing them from solving the task. Up to $V=10$, agents use the full available vocabulary, such that $V^{\,95\%}_{\mathrm{used}} = V$. However, with $V=2$, agents achieve only $19.84\%$ success, and with $V=10$, they achieve only moderate success ($76.09\%$). At these vocabulary sizes, the languages are highly conventional ($>90\%$), but the agents lack sufficient symbols to express the distinctions required by the task. Once the available vocabulary $V$ is sufficiently large, the population consistently converges to a compact convention across the tested vocabulary sizes. For $V \geq 35$, agents maintain high success ($>96\%$) and high conventionality ($>81\%$), while consistently converging to an effective vocabulary of 31 to 35 symbols. 

The Speaker-Listener baseline also suffers from under-capacity when $V$ is too small ($V \in \{2, 5, 10\}$), but it becomes increasingly sensitive to larger vocabulary sizes with respect to communicative success and conventionality. Larger vocabularies improve task success ($>84\%$), but they sharply decrease population-level alignment. Conventionality drops from $80.51\%$ at $V=10$ to $18.46\%$ at $V=100$, and to only $6.14\%$ at $V=500$. At the same time, the effective vocabulary size ($V^{\,95\%}_{\mathrm{used}}$) increases from 30 symbols at $V=35$ to 64 symbols at $V=500$. Thus, whereas our model stabilises around a compact effective vocabulary once sufficient capacity is available, the S-L baseline increasingly exploits the larger symbol space to the detriment of the emergent convention.

\begin{table*}[t]
\centering
\small
\begin{tabular}{cccc}
\toprule
\textbf{}                       & $\beta_i$ & Success & Conventionality \\
\midrule
\multirow{3}{*}{Bi-directional}   & 0 (No-Imitation) & 98.63$_{\pm 0.28}$       & 68.42$_{\pm 1.8}$ \\
                                  & 0.2 (Ours)       & 97.41$_{\pm 0.66}$       & 82.74$_{\pm 1.17}$ \\
                                  & 1                & 89.25$_{\pm 1.38}$       & 87.3$_{\pm 1.09}$ \\ 
\midrule
\multirow{3}{*}{Speaker-Listener} & 0                & 88.54$_{\pm 1.46}$       & 13.83$_{\pm 2.7}$ \\
                                  & 0.2              & 86.14$_{\pm 1.01}$       & 18.94$_{\pm 2.25}$ \\
                                  & 1                & 76.62$_{\pm 2.22}$       & 33.83$_{\pm 5.8}$ \\
\bottomrule  
\end{tabular}%
\caption{Results of the Bi-directional and Speaker-Listener models with different levels of imitation, controlled by $\beta_i$. Models were trained on CelebA for 2M interactions. Results are averaged over 10 independent runs and reported as mean $\pm$ standard deviation. }
\label{tab:SL+Im}
\end{table*}

The results show that the S-L baseline is sensitive to a task-dependent parameter: the vocabulary size $V$ must be large enough to solve the task, but small enough to maintain pressure towards conventionality. Crucially, the optimal vocabulary size cannot be determined a priori, as the number of required concepts to solve the task depends on the dataset and the number of entities in the scene. In contrast, our model consistently converges to a compact effective vocabulary once sufficient capacity is available.

\subsection{Speaker-Listener with imitation}
\label{app:SL+imitation}


To fully compare our model to the S-L baseline, we provide results of S-L when trained with the imitation loss. Table~\ref{tab:SL+Im} shows the results for three different values of the imitation loss coefficient $\beta_i\in\{0, 0.2, 1\}$, comparing the bi-directional and S-L models trained on CelebA for 2M interactions. In both cases, we observe a similar trade-off between success and conventionality controlled by the weight of imitation: increasing $\beta_i$ improves conventionality but also hurts success. While conventionality can be improved by imitation in S-L, the gain is not substantial. This demonstrates that imitation is not sufficient to provide a pressure for conventionality and that bi-directionality is a crucial feature of our model.

\subsection{Construction of Figure~\ref{fig:usage_top_100}}
\label{app:usage_details}
We describe here the exact procedure used to construct Figure~\ref{fig:usage_top_100}. The figure compares the relative usage frequency of symbols across agents in the \textit{CelebA} setup with 10 agents and a vocabulary size of 100. We compare our model against the S-L baseline, where 5 agents acted as speakers and 5 as listeners. Both models were trained until reaching 95\% communicative success, requiring approximately 500K interactions for our model and 20M for the S-L baseline. For each model, once training reached 95\% success, training was stopped and the population was evaluated for an additional 100K interactions on the evaluation set of unseen images, during which we recorded the symbols that were produced. For readability, we consider the first five agents of our model (S-L having only five speaking agents), but the pattern of symbol usage is similar across the whole population. For every plotted agent, we count how often each symbol was produced in successful games. Counts are normalised independently for each agent such that every column represents the agent's relative distribution over produced symbols. Within each model, symbols are ordered by their total frequency across the plotted agents, with the most frequently used symbols appearing at the top. 

\begin{figure}[t]
    \centering
    \includegraphics[width=\linewidth]{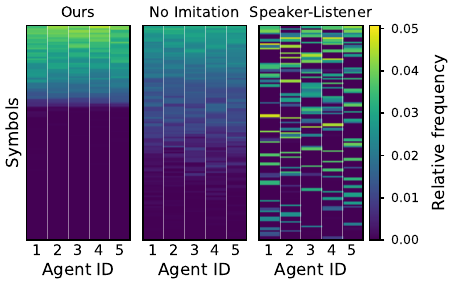}
    \caption{Relative usage frequency of symbols (rows) across 5 agents (columns) in the \textit{CelebA} setup. Comparing our model (Ours) with the No-Imitation and Speaker-Listener baselines. All three variants are trained until reaching 95\% success, but only our model produces highly aligned symbol usage across agents.}
    \label{fig:appendix-usage_top_100}
\end{figure}

Finally, Figure~\ref{fig:appendix-usage_top_100} provides an extended version of Figure~\ref{fig:usage_top_100} including the No-Imitation baseline. The No-Imitation population was also trained until it reached 95\% communicative success, which occurred after approximately 400K interactions. While the No-Imitation population exhibits partially overlapping symbol usage distributions, the degree of alignment remains substantially weaker than in our model.

\subsection{Scaling Populations}
\label{app:pop}

Training larger populations inevitably requires conducting more interactions because each agent has a lower chance of being selected in every interaction, and thus of gaining experience. For a population of $N$ agents, because each interaction selects a pair of agents, agents have a $2/N$ chance of being selected. Thus, we compare populations of different sizes with the same average number of interactions per agent. 

\begin{figure}[t]
    \centering
    \includegraphics[width=\linewidth]{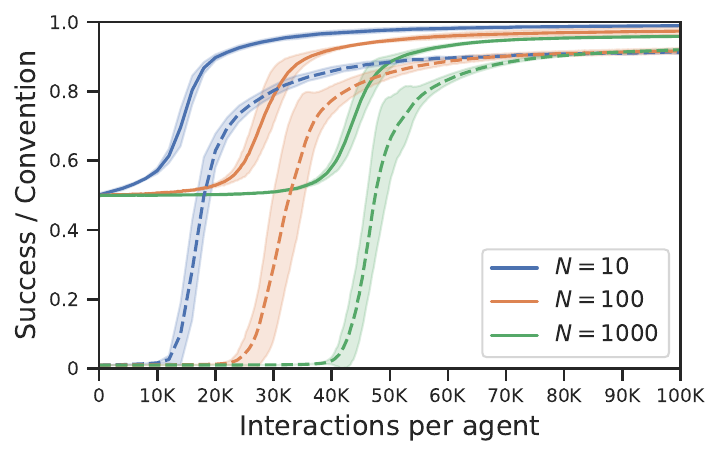}
    \caption{Training dynamics for populations of 10, 100 and 1000 agents on ManyNames. Curves show mean performance over 10 independent runs, with shaded regions indicating standard deviation. Solid lines indicate success, while dashed lines indicate conventionality.}
    \label{fig:pop}
\end{figure}

Table~\ref{tab:pop} shows the number of interactions done for populations of 10, 100 and 1000 agents, and their performance on ManyNames after 100K interactions per agent. Figure~\ref{fig:pop} also shows the performance throughout training. It clearly shows the delayed convergence for larger populations. Because more agents participate in the bootstrapping of the emergent language, it takes more time for successful behaviours to be transmitted in the population and finally adopted by everyone. However, note that all populations seem to converge to the same levels of success and conventionality, showing that the model works well with large populations. 

\begin{table}[t]
    \small
    \centering
    \begin{tabular}{cccc}
        \toprule
        $N$ & Total interactions & Success & Conv. \\
        \midrule
        10   & $5\times10^5$ & 98.97$_{\pm 0.19}$ & 91.23$_{\pm 0.50}$ \\
        100  & $5\times10^6$ & 97.40$_{\pm 1.11}$ & 91.62$_{\pm 0.80}$ \\
        1000 & $5\times10^7$ & 95.89$_{\pm 0.16}$ & 92.01$_{\pm 0.28}$ \\
        \bottomrule
    \end{tabular}
    \caption{Results of scaling the population size after training for an equivalent of 100K interactions per agent on ManyNames ($C=2$, $O=512$). Results are averaged over 10 independent runs and reported as mean $\pm$ standard deviation.}
    \label{tab:pop}
\end{table}

\subsection{Batched Interactions}
\label{app:batch}


In the batch experiment of Section~\ref{sec:batch}, we compare our model, which does not use batched interactions, with variants that use batch sizes of 10, 100 and 500. Increasing the batch size usually allows the learning rate to be increased, since  the computed updates are more stable. Thus, before the experiment, we performed a parameter search for the learning rate. For each batch size $B$ in $\{1\text{ (ours)},10,100,500\}$ we tested the values $\{0.00008, 0.0001, 0.00015,0.0002,0.0003,$ $0.0005,0.0008,0.001,0.0015,0.002\}$. The learning rates that gave the best trade-off between success and conventionality were: for $B=1$, $\alpha=0.0001$; for $B=10$, $\alpha=0.0002$; for $B=100$, $\alpha=0.0005$; and for $B=500$, $\alpha=0.0008$. All ``paired'' baselines are trained with the same learning rate as ours, $\alpha=0.0001$, because their agents are trained after each interaction.

\subsection{Symbol Usage Analysis}
\label{app:word_usage}

\paragraph{Construction of Figure~\ref{fig:word_usage}.}

\begin{figure}
    \centering
    \includegraphics[width=\linewidth]{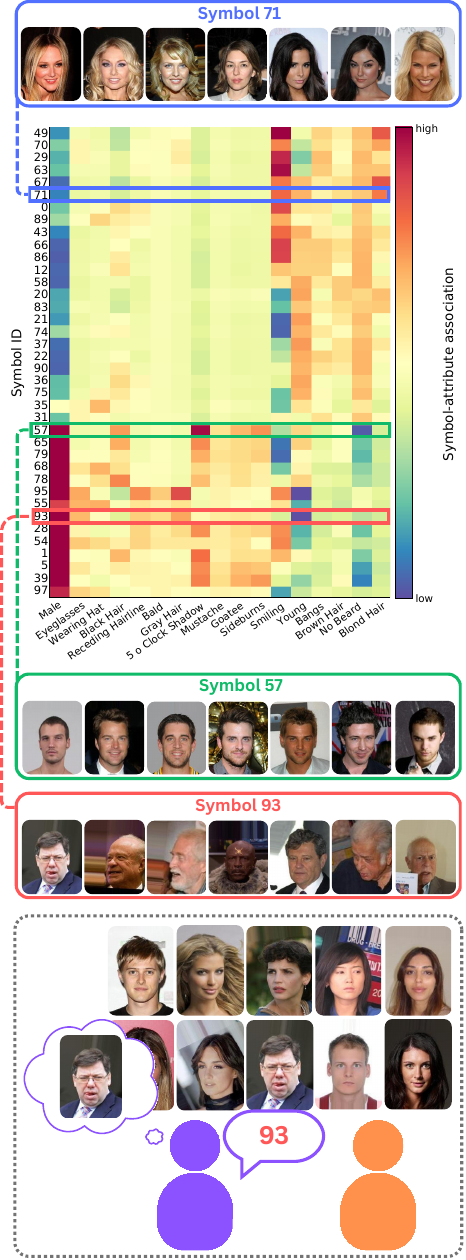}
    \caption{Extended version of Figure~\ref{fig:word_usage} from the main text showing co-occurrences between symbols and \textit{CelebA} attributes for all 38 ``active'' symbols that were produced at least 50 times during 100K evaluation interactions on unseen images. Three symbols are highlighted with example topic images that elicited those symbols. At the bottom, an example interaction is shown in which symbol 93 was successfully used to discriminate the topic image from the distractor set.}
    \label{fig:appendix-assoc}
\end{figure}

In Section~\ref{sec:word_usage} of the main text, we introduced Figure~\ref{fig:word_usage}. We describe here the exact procedure used to construct the symbol-attribute association heatmap and the accompanying example images. The figure was produced for our model in the \textit{CelebA} setup with 10 agents and a vocabulary size of 100. The population was trained for 2M interactions. After training, the population was evaluated on an additional 100K interactions using the evaluation set of unseen images. During these interactions, we recorded the produced symbols together with the topic image and the distractor set shown in each game.

The \textit{CelebA} dataset contains 40 binary visual attributes annotated by humans for each image. These annotations describe visual properties such as whether the person is smiling, bald, blonde-haired, etc. Importantly, these attribute annotations are never observed by agents during training or evaluation and play no role in the learning process itself. They are used only afterwards as a post hoc analysis tool to interpret the semantic structure of the emerged symbols. Consequently, this analysis is restricted to the subset of properties explicitly annotated in the dataset metadata. The learned symbols may therefore also encode other latent visual regularities that are not captured by these annotations, such as background colour, head orientation or lighting conditions.

For every produced symbol $s$ and every \textit{CelebA} attribute $a$, we computed a contrastive association score measuring how much more frequently the topic image possessed the attribute compared to the distractor images in interactions where the symbol was produced. Positive values indicate that the symbol is produced more frequently when the topic image possesses the attribute than when the distractors do, whereas negative values indicate the opposite. 

Due to space constraints and visualisation considerations, Figure~\ref{fig:word_usage} restricts the heatmap to the 15 most frequently produced symbols and the 17 comparatively salient and easily identifiable attributes, including facial hair, hair colour, accessories or age. Some of the excluded attributes are considerably more difficult to assess consistently from visual inspection alone, such as ``\textit{pointy nose}''. Each row corresponds to a symbol ID and each column to a \textit{CelebA} attribute. Each cell visualises the corresponding contrastive association score. The colour scale is centred around zero, with warmer colours corresponding to positive associations and colder colours to negative associations. Rows and columns were independently ordered using hierarchical clustering of the association scores using Euclidean distance. This clustering procedure was used solely for visualisation purposes to group similar rows and columns.

For symbols 71 and 93, we additionally  display three example topic images sampled from evaluation interactions in which the corresponding symbol was produced by an agent. For each symbol, we ranked all corresponding interactions using the model's top-symbol discriminative score. The examples correspond to concrete cases in which the symbol was successfully used to discriminate the topic from a particular set of distractors. We highlighted the three highest-ranking examples for each symbol.

Finally, Figure~\ref{fig:appendix-assoc} provides a full version of the heatmap containing all 38 active symbols that were produced at least 50 times during the 100K evaluation interactions. In addition to the original examples, we highlight a third symbol (57), which is associated with younger males with light facial hair. For the three highlighted symbols, we display the seven highest-ranking examples rather than only the top three. These extended examples further illustrate that the learned symbols capture broad bundles of correlated properties. Finally, at the bottom of Figure~\ref{fig:appendix-assoc}, we provide an example interaction in which symbol 93 was successfully used. As shown, the symbol is used to discriminate the older gentleman (i.e. the topic of the interaction) from the relatively younger individuals in the distractor set.



\section{AI Assistant Use}

All authors declare that this article was conceived and written without the use of generative AI tools, except to help write the code for generating the figures. They were \textit{not} used to generate code used in the experiments.


\newpage
\onecolumn

\section{Derivation of the Training Objectives}
\label{app:DiscrScore}

\subsection{Speaker objective}

For a speaker, the objective is to maximise the success of the words it utters to describe entities. Here, we detail how the speaker, denoted $S$, computes its utterance distribution and how optimising this objective will impact the learnt representations. 

The two networks to optimise are the entity encoder and the word encoder, with parameters $\theta_{S,e}$ and $\theta_{S,w}$, respectively. They produce the embeddings for observed entities and words in the vocabulary, $f(e_i|\theta_{S,e})$ and $f(w_j|\theta_{S,w})$, which will determine which word can be used for which entity. To this end, the basic component is the similarity score computed between these embeddings, defined as the scaled dot product: 
\begin{equation}
    \Sim(e,w|\theta_{S})=\frac{f(e|\theta_{S,e})\cdot f(w|\theta_{S,w})^\top}{\sqrt{E}},
\end{equation}
with $\theta_S=\{\theta_{S,e},\theta_{S,w}\}$ and $E$ the dimension of the embeddings. 
Given a scene $e=\{e_1,...,e_C\}$ and the topic $e_t\in e$, the \textbf{discrimination score} of any word $w_j$ can be computed from the similarity scores:
\begin{equation}
    \delta(e_t,w_j|\theta_S)=\Sim(e_t,w_j|\theta_S)-\max_{i\neq t}\Sim(e_i,w_j|\theta_S).
\end{equation}
This score measures the discriminative power of each word in the vocabulary $w_j$ for the current scene and topic, by taking the similarity between $w_j$ and the topic $e_t$, and subtracting the highest similarity between $w_j$ and other entities in the scene. 
To allow this score to be fully differentiable with regard to all other entities in the scene, we can replace the $\max$ function by the differentiable estimate $LogSumExp$:
\begin{align}
    \delta(e_t,w_j|\theta_S) &= \Sim(e_t,w_j|\theta_S) - LogSumExp_{i\neq t}\bigl(\Sim(e_i,w_j|\theta_S)/\tau_{lse}\bigr)\\
                           &= \Sim(e_t,w_j|\theta_S) - \log\sum_{i\neq t}e^{\Sim(e_i,w_j|\theta_S)/\tau_{lse}},
\end{align}
with $\tau_{lse}$ a temperature parameter that can be used to make $LogSumExp$ closer to the $\max$ function (in practice we keep it at 1). 
The speaker's policy is the distribution over words in the vocabulary derived from the discrimination score:
\begin{align}
    \pi(w_j|e,e_t,\theta_S) &= softmax\bigl(\delta(e_t, w|\theta_S) / \tau_{s}\bigr)_j\\
                      &= \frac{e^{\delta(e_t, w_j|\theta_S)/\tau_s}}{\sum_{k}e^{\delta(e_t, w_k|\theta_S)/\tau_{s}}},
\end{align}
with $\tau_{s}$ a temperature parameter controlling the flatness of this distribution, which can be used to control the level of exploration in the speaker's policy. 

In each interaction, the speaker samples a word $w_S$ from this distribution. The outcome of the interaction gives a reward $R$ equal to either 1 for success or 0 for failure. From this reward, we compute the speaker's learning objective as the REINFORCE loss:
\begin{equation}
    \mathcal{L}_{speak}(\theta_S)=-\log\bigl(\pi(w_S|e,e_t,\theta_S)\bigr)R.
\end{equation}
Minimising this loss will increase the probability of selecting words that have led to success. 
To understand how this optimisation process impacts the word representations, we can compute the gradients of this loss with respect to the speaker's parameters to determine how the similarity scores are updated:
\begin{gather}
    \nabla_{\theta_S}\mathcal{L}_{speak}(\theta_S)=-R\nabla_{\theta_S}\log\pi(w_S|e,e_t,\theta_S).
\end{gather}
Let $z_j=\frac{\delta(e_t,w_j|\theta_S)}{\tau_s}$, for each word in the vocabulary $w_j$,
\begin{align}
    \nabla_{\theta_S}\mathcal{L}_{speak}(\theta_S) &= -R\frac{\partial}{\partial z_j}\log\pi(w_S|e,e_t,\theta_S)\nabla_{\theta_S}z_j\\
                                 &= -R\bigl(\mathds{1}\{j=S\}-\pi(w_j|e,e_t,\theta_S)\bigr)\nabla_{\theta_S}z_j\\
                                 &= R\bigl(\pi(w_j|e,e_t,\theta_S)-\mathds{1}\{j=S\}\bigr)\nabla_{\theta_S}z_j.
\end{align}
Let $K=R\bigl(\pi(w_j|e,e_t,\theta_S)-\mathds{1}\{j=S\}\bigr)$,
\begin{align}
    \nabla_{\theta_S}\mathcal{L}_{speak}(\theta_S) &= K\nabla_{\theta_S}z_j\\
                                 &= \frac{K}{\tau_s}\nabla_{\theta_S}\delta(e_t,w_j|\theta_S)\\
                                 &= \frac{K}{\tau_s}\nabla_{\theta_S}\Bigl[\Sim(e_t,w_j|\theta_S) - \log\sum_{i\neq t}e^{\Sim(e_i,w_j|\theta_S)/\tau_{lse}}\Bigr]\\
                                 &= \underbrace{\frac{K}{\tau_s}\nabla_{\theta_S}\Sim(e_t,w_j|\theta_S)}_{D_t} - \underbrace{\frac{K}{\tau_s}\nabla_{\theta_S}\log\sum_{i\neq t}e^{\Sim(e_i,w_j|\theta_S)/\tau_{lse}}}_{D_{t^-}}
\end{align}
\begin{align}
    D_{t^-} &= -\frac{K}{\tau_s}\nabla_{\theta_S}\log\sum_{i\neq t}e^{\Sim(e_i,w_j|\theta_S)/\tau_{lse}}\\
            &= -\frac{K}{\tau_s}\frac{1}{\sum_{i\neq t}e^{\Sim(e_i,w_j|\theta_S)/\tau_{lse}}}\sum_{i\neq t}e^{\Sim(e_i,w_j|\theta_S)/\tau_{lse}}\frac{1}{\tau_{lse}}\nabla_{\theta_S}\Sim(e_i,w_j|\theta_S) \\
            &= -\frac{K}{\tau_s\tau_{lse}}\sum_{i\neq t}\left[\frac{e^{\Sim(e_i,w_j|\theta_S)/\tau_{lse}}}{\sum_{k\neq t}e^{\Sim(e_k,w_j|\theta_S)/\tau_{lse}}}\nabla_{\theta_S}\Sim(e_i,w_j|\theta_S)\right]\\
            &= -\frac{K}{\tau_s\tau_{lse}}\sum_{i\neq t}\Bigl[softmax\bigl(\Sim(e_{\neq t},w_j|\theta_S)/\tau_{lse}\bigl)_i\nabla_{\theta_S}\Sim(e_i,w_j|\theta_S)\Bigr]
\end{align}
The terms $D_t$ and $D_{t^-}$ describe how the similarities between observed entities in the scene and words in the speaker's vocabulary are updated. $D_t$ concerns the similarities between the topic and all words, and $D_{t^-}$ the similarities between all the distractors and all words. 

Because the reward $R$ is $0$ in failure cases, $K=0$ and there is no update in these cases. In other words, the speaker will not change its representations if the interaction has failed. This shows an (intended) bias towards the speaker's choices in this algorithm: when a speaker chooses a word to utter, this choice is considered valid (while not necessarily perfect) and the listener should try to understand it (and optimise for it). 

In the successful case, $K=\pi_S(w_j)-\mathds{1}\{j=S\}$ (for the purpose of explanation, we note $\pi(w_j|e,e_t,\theta_S)$ as $\pi_S(w_j)$), which gives the following updates:
\begin{itemize}
    \item For $j=S$, i.e. the word uttered by the speaker, we have $K=\pi_S(w_S)-1$. Because $\pi_S(w_S)\in[0,1]$, then $K\leq0$. Thus, minimising $D_t$ will increase $\Sim(e_t,w_S|\theta_S)$ proportionally to $\frac{|\pi_S(w_S)-1|}{\tau_s\tau_{lse}}$. \\
    For $D_{t^-}$, $-K=1-\pi_S(w_S)\geq0$, thus minimising $D_{t^-}$ will decrease each $\Sim(e_i,w_S|\theta_S)$, with $i\neq t$, proportionally to $\frac{1-\pi_S(w_S)}{\tau_s\tau_{lse}}softmax\bigl(\Sim(e,w_S)/\tau_{lse}\bigl)_i$. \\
    In other words, the update increases the similarity between the embeddings of the topic entity and the uttered word proportionally to the output probability of uttering this word (\textbf{reward successful word}), and decreases the similarities between the distractor entities and the uttered word proportionally to their uttering probability and to their relative weight in the similarity score (\textbf{punishing competitors}). 

    \item For $j\neq S$, i.e. all the other words not chosen by the speaker, $K=\pi_S(w_j)\geq0$. Minimising $D_t$ will decrease the similarities $\Sim(e_t,w_j|\theta_S)$, proportionally to $\frac{\pi_S(w_j)}{\tau_s\tau_{lse}}$. \\
    Minimising $D_{t^-}$ with $-K=-\pi_S(w_j)\leq0$ will increase the similarities $\Sim(e_i,w_j|\theta_S)$, with $i\neq t$, proportionally to $\frac{\pi_S(w_j)}{\tau_s\tau_{lse}}softmax\bigl(\Sim(e,w_j)/\tau_{lse}\bigl)_i$. \\
    In other words, this update decreases the similarity between the words that were not uttered and the topic entity proportionally to their uttering probability (\textbf{punishing competitors}), and increases the similarity between the words not uttered and the distractor entities proportionally to their uttering probability and to their relative weight in the similarity score (\textbf{realigning representations}). 
\end{itemize}

\subsection{Listener objective}

As a listener, the objective is to correctly understand the speaker's utterance and to align its speaking policy with the speaker. In each interaction, the listener, denoted $L$, observes the scene $e=\{e_1,...,e_C\}$ and the speaker's utterance $w_S$. It computes a distribution over the entities in the scene from which it samples its prediction (in practice, we always take the entity with the highest predicted probability). The distribution is computed directly from the similarity scores between the received word and each entity in the scene:
\begin{equation}
    p_L(e_i|e, w_S, \theta_L)=softmax\bigl(\Sim(e,w_S|\theta_L)\bigr)_i.
\end{equation}
The listener's first learning objective is to maximise the probability given to the topic, i.e. minimise the cross-entropy loss:
\begin{equation}
    \mathcal{L}_{pred}(\theta_L) = -\log\bigl(p_L(e_t|e,w_S,\theta_L)\bigr).
\end{equation}
With a computation similar to that for the speaker loss, this gives the following gradients:
\begin{equation}
    \nabla_{\theta_L}\mathcal{L}_{pred}(\theta_L) = \bigl(p_L(e_i|e,w_S,\theta_L)-\mathds{1}\{i=t\}\bigr)\nabla_{\theta_L}\Sim(e_i,w_S|\theta_L),
    \label{eq:Lpred}
\end{equation}
for each entity $e_i$ in the scene. 

A second learning objective is optimised by the listener to align its word usage with the speaker. This is done by updating the listener's speaking policy towards using the same word as the speaker to describe the topic in this scene. The probability of the speaker's word $w_S$ in the listener's policy is:
\begin{equation}
    \pi(w_S|e,e_t,\theta_L)=softmax\bigl(\delta(e_t, w|\theta_L)\bigr)_S.
\end{equation}
The imitation objective maximises this probability, by minimising the following loss:
\begin{equation}
    \mathcal{L}_{imit}(\theta_L) = -\log\pi(w_S|e,e_t,\theta_L).
\end{equation}
This is almost the same objective as the speaker's $\mathcal{L}_{speak}$, without the reward $R$ and with all temperature parameters set to 1. Thus, we can derive the gradients of $\mathcal{L}_{imit}$, for each word $w_j$, as:
\begin{align}
\begin{split}
    \nabla_{\theta_L}\mathcal{L}_{imit}(\theta_L) &= K\bigl(D_t-D_{t^-}\bigr), \\
                         \text{with } K &= \bigl(\pi(w_j|e,e_t,\theta_L)-\mathds{1}\{j=S\}\bigr), \\
                                    D_t &= \nabla_{\theta_L}\Sim(e_t,w_j|\theta_L), \\
                    \text{and } D_{t^-} &= \sum_{i\neq t}\Bigl[softmax\bigl(\Sim(e_{\neq t},w_j|\theta_L)\bigl)_i\nabla_{\theta_L}\Sim(e_i,w_j|\theta_L)\Bigl].
\end{split}
\label{eq:Limit}
\end{align}

From these equations, we can understand how minimising $\mathcal{L}_{pred}$ and $\mathcal{L}_{imit}$ aligns the listener's policy to the speaker's. Equation \ref{eq:Lpred} shows that:
\begin{itemize}
    \item For $i=t$, i.e. the topic, $p_L(e_t) - 1 \leq 0$, so minimising the objective will increase the similarity between the topic and the speaker's utterance.
    \item For $i\neq t$, i.e. all other entities in the scene, $p_L(e_i) \geq 0$, so the similarity between the distractor entities and the speaker's utterance will be decreased. 
\end{itemize}
Equation \ref{eq:Limit} mirrors the update of the speaker:
\begin{itemize}
    \item For $j=S$, i.e. the word uttered by the speaker, we have $K=\pi_L(w_S)-1 \leq0$, thus minimising $D_t$ will increase the similarity between the topic and the speaker's word. \\
    For $D_{t^-}$, $-K=1-\pi_L(w_S)\geq0$, thus minimising $D_{t^-}$ will decrease the similarities between the distractor entities and the speaker's word.
    
    \item For $j\neq S$, i.e. all the other words not chosen by the speaker, $K=\pi_L(w_j)\geq0$, thus minimising $D_t$ will decrease the similarities between the topic and the words. \\
    Minimising $D_{t^-}$ with $-K=-\pi_L(w_j)\leq0$ will increase the similarities between the distractor entities and the other words in the vocabulary. 
\end{itemize}

\end{document}